\documentclass[final,12pt]{clear2022} % Include author names

\usepackage{hyperref} 
\hypersetup{colorlinks,linkcolor={black},citecolor={[rgb]{0.46875   , 0.57421875, 0.58203125}},urlcolor={[rgb]{0.578125  , 0.703125  , 0.62109375}}} 
\usepackage{algorithm}
\usepackage{algpseudocode}
\usepackage{caption}
\usepackage{amsmath}

\usepackage{wrapfig}
\usepackage{amsmath}
\usepackage{mathtools}
\usepackage{xspace}

\usepackage[most]{tcolorbox}
\usepackage{sidecap}
\usepackage{booktabs}

\usepackage[inline]{enumitem}
\newlist{choices}{enumerate*}{1}
\setlist[choices]{itemsep = 1.125in, label=(\roman*)}

\usepackage{amsmath,amsfonts,bm}

\def\eqref#1{equation~\ref{#1}}
\def\1{\bm{1}}

\def\mI{{\bm{I}}}

\DeclareMathAlphabet{\mathsfit}{\encodingdefault}{\sfdefault}{m}{sl}
\SetMathAlphabet{\mathsfit}{bold}{\encodingdefault}{\sfdefault}{bx}{n}

\def\gN{{\mathcal{N}}}

\newcommand{\R}{\mathbb{R}}

\newcommand{\norm}[1]{\left\lVert#1\right\rVert}

\DeclarePairedDelimiterX{\infdivx}[2]{(}{)}{%
  #1\;\delimsize|\delimsize|\;#2%
}
\newcommand{\kld}[2]{\ensuremath{D_{KL}\infdivx{#1}{#2}}\xspace}

\title[Diff-SCM for Counterfactual Estimation]{Diffusion Causal Models for Counterfactual Estimation}

\usepackage{times}

\clearauthor{\Name{Pedro Sanchez}  \Email{pedro.sanchez@ed.ac.uk} \\
\Name{Sotirios A. Tsaftaris}  \\
\addr The University of Edinburgh
}
\begin{document}

\maketitle

\begin{figure}[ht]
\centering
\includegraphics[width=\linewidth]{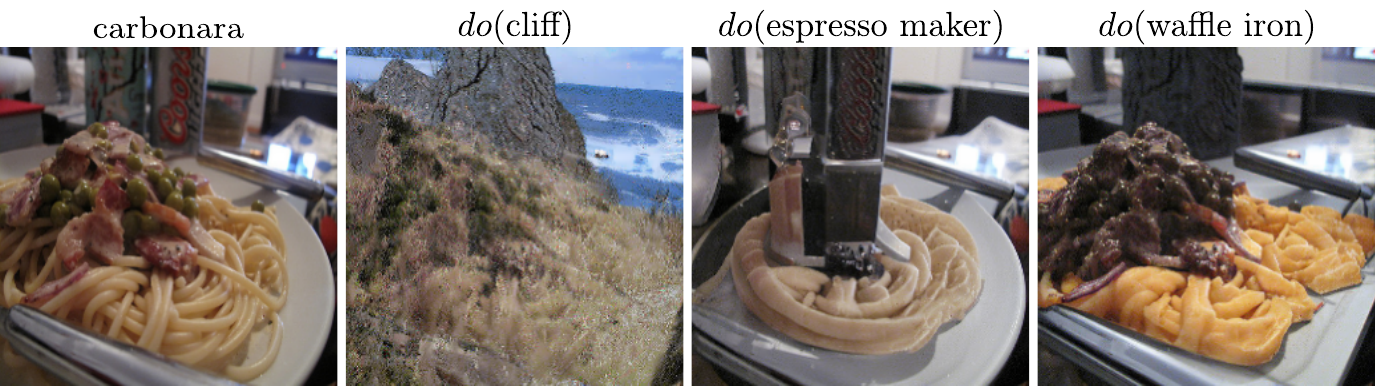}
\caption{Counterfactuals on ImageNet 256x256 generated by Diff-SCM. \textit{From left to right}: a random image sampled from the data distribution and its counterfactuals $do(\text{class})$, corresponding to ``how the image should change in order to be classified as another class?''.}
\label{fig:qualitative_ImageNet_counterfactuals_samples}
\end{figure}

\begin{abstract}

We consider the task of counterfactual estimation from observational imaging data given a known causal structure. In particular, quantifying the causal effect of interventions for high-dimensional data with neural networks remains an open challenge. Herein we propose Diff-SCM, a deep structural causal model that builds on recent advances of generative energy-based models. In our setting, inference is performed by iteratively sampling gradients of the marginal and conditional distributions entailed by the causal model. Counterfactual estimation is achieved by firstly inferring latent variables with deterministic forward diffusion, then intervening on a reverse diffusion process using the gradients of an anti-causal predictor w.r.t the input. Furthermore, we propose a metric for evaluating the generated counterfactuals. We find that Diff-SCM produces more realistic and minimal counterfactuals than baselines on MNIST data and can also be applied to ImageNet data.
Code is available \url{https://github.com/vios-s/Diff-SCM}.

\end{abstract}
\section{Introduction}

The notion of applying interventions in learned systems has been gaining significant attention in causal representation learning \citep{Scholkopf2021TowardLearning}. In causal inference, relationships between variables are directed. An intervention on the cause will change the effect, but not the other way around. This notion goes beyond learning conditional distributions $p(\mathbf{x}^{(k)} \mid \mathbf{x}^{(j)})$ based on the data alone, as in the classical statistical learning framework \citep{Vapnik1999AnTheory}. Building causal models implies capturing the underlying physical mechanism that generated the data into a model \citep{Pearl2009causality}. As a result, one should be able to quantify the causal effect of a given action. In particular, when an intervention is applied for a given instance, the model should be able the generate hypothetical scenarios. These are the so-called \textit{counterfactuals}.

Building causal models that quantify the effect of a given action for a given causal structure and available data is referred to as \textit{causal estimation}. However, estimating the effect of interventions for high-dimensional data remains an open problem \citep{Pawlowski2020DeepInference, Yang2021CausalVAE:Models}. While machine learning is a powerful tool for learning relationships between high-dimensional variables, most causal estimation methods using neural networks \citep{Johansson2016LearningInference,Louizos2017CausalModels,Shi2019AdaptingEffects,Du2021AdversarialData} are only applied in semi-synthetic low-dimensional datasets \citep{Hill2012BayesianInference,Shimoni2018BenchmarkingAnalysis}. Therefore, causal estimation through learning deep neural networks for high-dimensional variables remains a desired quest. We show that we can estimate the effect of interventions by generating counterfactuals on imaging datasets, as illustrated in Fig.\ \ref{fig:qualitative_ImageNet_counterfactuals_samples}.

%% MY SOLUTION
Herein, we leverage recent advances in generative energy based models (EBMs) \citep{Song2021Score-BasedEquations,Ho2020DenoisingModels} to devise approaches for causal estimation. This formulation has two key advantages: (i) the stochasticity of the diffusion process relates to uncertainty-aware causal models; and (ii) the iterative sampling can be naturally extended for applying interventions. Additionally, we propose an algorithm for counterfactual inference and a metric for evaluating the results. In particular, we use neural networks that learn to reverse a diffusion process \citep{Ho2020DenoisingModels} via denoising. These models are trained to approximate the gradient of a log-likelihood of a distribution w.r.t. the input. We also employ neural networks that are learned in the anti-causal direction \citep{Scholkopf2012OnLearning,Kilbertus2018GeneralizationLearning} to sample via the causal mechanisms. We use the gradients of these anti-causal predictors for applying interventions in specific variables during sampling. Counterfactual estimation is possible via a deterministic version of diffusion models \citep{Song2021DenoisingModels} which recovers manipulable latent spaces from observations. Finally, the counterfactuals are generated iteratively using Markov Chain Monte Carlo (MCMC) algorithms.

In summary, we devise a framework for causal effect estimation with high-dimensional variables based on diffusion models entitled Diff-SCM. Diff-SCM behaves as a structured generative model where one can sample from the interventional distribution as well as estimate counterfactuals. Our contributions:
\begin{choices}
    \item We propose a theoretical framework for causal modeling using generative diffusion models and anti-causal predictors (Sec.\ \ref{sec:causal_models_diffusion}).
    \item We investigate how anti-causal predictors can be used for applying interventions in the causal direction (Sec.\ \ref{sec:anti_causal_interventions}).
    \item We propose an algorithm for counterfactual estimation using Diff-SCM (Sec.\ \ref{sec:counterfactuals}).
    \item We propose a metric term counterfactual latent divergence for evaluating the \textit{minimality} of the generated counterfactuals (Sec.\ \ref{sec:cld}). We use this metric to compared our method with the selected baselines and hyperparameter search (Sec.\ \ref{sec:hyper_search})
\end{choices}

\section{Background}
\subsection{Generative Energy-Based Models}
\label{sec:energy_based_models}
A family of generative models based on diffusion processes \citep{Sohl-Dickstein2015DeepThermodynamics,Ho2020DenoisingModels, Song2021Score-BasedEquations} has recently gained attention even achieving state-of-the-art image generation quality \citep{Dhariwal2021DiffusionSynthesis}.

In particular, Denoising Diffusion Probabilistic Models (DDPMs) \citep{Ho2020DenoisingModels} consist in learning to denoise images that were corrupted with Gaussian noise at different scales. DDPMs are defined in terms of a forward Markovian diffusion process. This process gradually adds Gaussian noise, with a time-dependent variance $\beta_t \in [0,1]$, to a data point $\mathbf{x}_0 \sim p_{\text{data}}(\mathbf{x})$. Thus, the latent variable $\mathbf{x}_t$, with $t \in \left[ 0,T\right]$, is learned to correspond to versions of $\mathbf{x}_0$ perturbed by Gaussian noise following 
$p \left(\mathbf{x}_t \mid \mathbf{x}_0\right)=\mathcal{N}\left(\mathbf{x}_t ; \sqrt{\alpha_{t}} \mathbf{x}_0,\left(1-\alpha_{t}\right) \mathrm{I}\right)$, where $\alpha_{t}:=\prod_{j=0}^{t}\left(1-\beta_{j}\right)$ and $\mathrm{I}$ is the identity matrix. 

As such, $p(\mathbf{x}_t) =\int p_{\text{data}}(\mathbf{x}) p(\mathbf{x}_t \mid \mathbf{x}) \mathrm{d} \mathbf{x} $ should approximate the data distribution $p(\mathbf{x}_0) \approx p_{\text{data}}$ at time $t = 0$ and a zero centered Gaussian distribution at time $t = T$. Generative modelling is achieved by learning to reverse this process using a neural network $\boldsymbol{\epsilon}_{\theta}$ trained to denoise images at different scales $\beta_t$. The denoising model effectively learns the gradient of a log-likelihood w.r.t. the observed variable $\nabla_{\mathbf{x}} \log p(\mathbf{x})$ \citep{Hyvarinen2005EstimationMatching}.

\textbf{Training.}~ With sufficient data and model capacity, the following training procedure ensures that the optimal solution to $\nabla_{\mathbf{x}} \log p_t(\mathbf{x})$ can be found by training $\boldsymbol{\epsilon}_{\theta}$ to approximate $\nabla_{\mathbf{x}} \log p_t(\mathbf{x}_t \mid \mathbf{x}_0)$. The training procedure can be formalised as
\begin{equation}
\label{eq:trainingDDPM}
\theta^{*}=\underset{\theta}{\arg \min }
~\mathbb{E}_{t,\mathbf{x}_0 \sim p_{\text{data}},\epsilon \sim \mathcal{N}\left(0, \mathrm{I}\right)} \left[ (1 - \alpha_t)
\left\| \boldsymbol{\epsilon}_{\theta}(\sqrt{\alpha_t}\mathbf{x}_0 + \sqrt{1 - \alpha_t} \epsilon, t)- \epsilon \right\|_{2}^{2}\right].
\end{equation}

\textbf{Inference.}~Once the model $\boldsymbol{\epsilon}_{\theta}$ is learned using Eq.\ \ref{eq:trainingDDPM}, generating samples consists in starting from $\mathbf{x}_T \sim \mathcal{N}(\mathbf{0}, \mathrm{I})$ and iteratively sampling from the reverse Markov chain following:
\begin{equation}
\label{eq:ddpm_sampling}
\mathbf{x}(t-1)=\frac{1}{\sqrt{1-\beta_{t}}}\left[\mathbf{x}_t+\beta_{t}~ \boldsymbol{\epsilon}_{\theta^{*}}(\mathbf{x}_t, t)\right]+\sqrt{\beta_{t}} \mathbf{z} , \quad t = T \cdots 0 ~,~ \mathbf{z} \sim \mathcal{N}(\mathbf{0}, \mathrm{I}).
\end{equation}
%With $\mathbf{z} \sim \mathcal{N}(\mathbf{0}, \mathrm{I})$.

We note that, in the DDPM setting, $\mathbf{z}$ is re-sampled at each iteration. Diffusion models are Markovian and stochastic by nature. As such, they can be defined as a stochastic differential equation (SDE) \citep{Song2021Score-BasedEquations}. We adopt the time-dependent notation from \citet{Song2021Score-BasedEquations} as it will be useful for the connection with causal models in Sec.\ \ref{sec:causal_models_diffusion}.

\subsection{Causal Models}
\label{sec:causal_models}
Counterfactuals can be understood from a formal perspective using the causal inference formalism \citep{Pearl2009causality,Peters2017ElementsInference,Scholkopf2021TowardLearning}. Structural Causal Models (SCM) $\mathfrak{G} := (\mathbf{S},p_{U})$ consist of a collection $\mathbf{S} = (f^{(1)},f^{(2)},....,f^{(K)})$  of structural assignments (so-called \textit{mechanisms}), defined as
\begin{equation}
\label{eq:structural_assignments}
    \mathbf{x}^{(k)} := f^{(k)}(\mathbf{pa}^{(k)},\mathbf{u}^{(k)}),
\end{equation}
where $X = \{ \mathbf{x}^{(1)},\mathbf{x}^{(2)},...,\mathbf{x}^{(K)}\}$ are the known endogenous random variables, $\mathbf{pa}^{(k)}$ is the set of parents of $\mathbf{x}^{(k)}$ (its direct causes) and $U = \{\mathbf{u}^{(1)},\mathbf{u}^{(2)},...,\mathbf{u}^{(K)}\}$ are the exogenous variables. The distribution $p(U)$ of the exogenous variables represents the uncertainty associated with variables that were not taken into account by the causal model. Moreover, variables in $U$ are  mutually independent following the joint distribution:
\begin{equation}
\label{eq:u_independence}
    p(U) = \prod_{k=1}^{K} p(\mathbf{u}^{(k)}).
\end{equation}

These structural equations can be defined graphically as a directed acyclic graph. Vertices are the endogenous variables and edges represent (directional) causal relationships between them. In particular, there is a joint distribution $ p_{\mathfrak{G}}(X) = \prod_{k = 1}^K p(\mathbf{x}^{(k)} \mid \mathbf{pa}^{(k)})$ which is Markov related to $\mathcal{G}$. In other words, the SCM $\mathfrak{G}$ represents a joint distributions over the endogenous variables. A graphical example of a SCM is depicted on the left part of Fig.\ \ref{fig:diffusion_causal_relations}. Finally, SCMs should comply to what is known as \textit{Pearl's Causal Hierarchy} (see Appendix \ref{app:pch} for more details).

\section{Causal Modeling with Diffusion Processes}

\subsection{Problem Statement}

In this work, we build a causal model capable of estimating counterfactuals of high-dimensional variables. We will base our work on three assumptions:
\begin{choices}
    \item The SCM is known and the intervention is identifiable.
    \item The variables over which the counterfactuals will be estimated need to contain enough information to recover their causes; \textit{i.e.}\ an anti-causal predictor can be trained. 
    \item All endogenous variables in the training set are annotated.
\end{choices}

\textbf{Notation.}~We use $\mathbf{x}^{(k)}_t$ is the $k^{th}$ endogenous random variable in a causal graph $\mathfrak{G}$ at diffusion time $t$. $x^{(k)}_{t,i}$ is a sample $i \in [\text{CF}, \text{F}]$ (F and CF being factual and counterfactual respectively) from $\mathbf{x}^{(k)}_t$. Whenever $t$ is omitted, it should be considered zero, \textit{i.e.}\ the sample is not corrupted with Gaussian noise.  $\mathbf{an}^{(k)}$ for the ancestors, with $\mathbf{pa}^{(k)} \subset \mathbf{an}^{(k)}$, and $\mathbf{de}^{(k)}$ for the descendants of $\mathbf{x}^{(k)}$ in $\mathfrak{G}$.

\subsection{Diff-SCM: Unifying Diffusion Processes and Causal Models}
\label{sec:causal_models_diffusion}

\begin{figure}[ht]
\centering
\includegraphics[width=\linewidth]{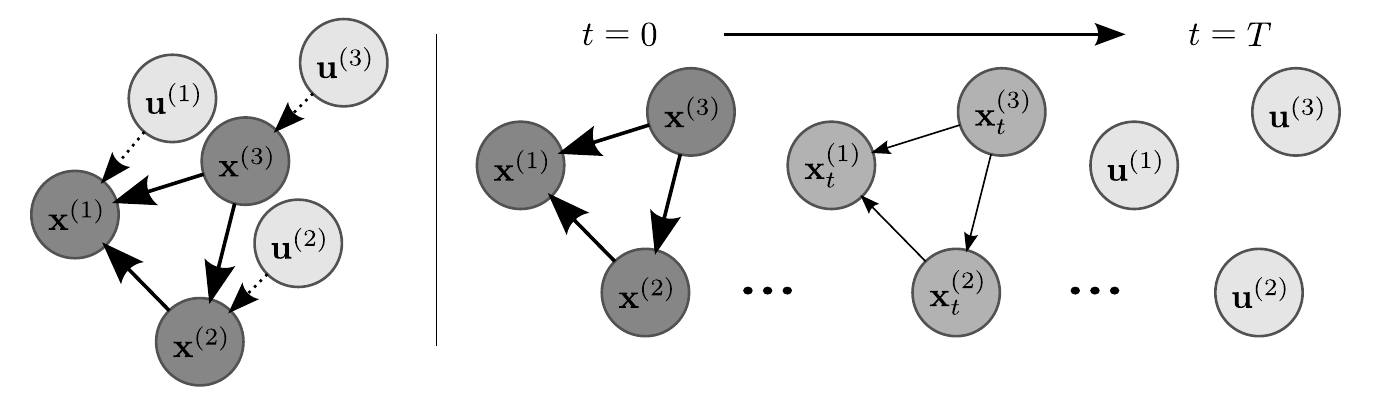}
\caption{Illustration of a diffusion process as weakening of causal relationships. \textit{Left:}~Example of a SCM with endogenous variables $\mathbf{x}^{(k)}$ and respective exogenous variables $\mathbf{u}^{(k)}$. \textit{Right:}~The diffusion process weakens the relationship between endogenous variables until they become completely independent at $t = T$. Arrows with solid lines indicate the causal relationship between variables and direction, while the thickness of the arrow indicates strength of the relation. Note that time $t$ is a fiction used as reference for the diffusion process and is \textit{not} a causal variable.}
\label{fig:diffusion_causal_relations}
\end{figure}

SCMs have been associated with ordinary \citep{Mooij2013FromCase,Rubenstein2018FromModels} and stochastic \citep{Sokol2014CausalEquations,Bongers2018FromCase} differential equations as well as other types of dynamical systems \citep{Blom2020BeyondModels}. In these cases, differential equations are useful for modeling time-dependent problems such as chemical kinetics or mass-spring systems. From the energy-based models perspective, \citet{Song2021Score-BasedEquations} unify denoising diffusion models \citep{Sohl-Dickstein2015DeepThermodynamics,Ho2020DenoisingModels} and denoising score models \citep{Song2019GenerativeDistribution} into a framework based on SDEs. In \citet{Song2021Score-BasedEquations}, SDEs are used for formalising a diffusion process in a continuous manner where a model is learned to reverse the SDE in order to generate images.

Here, we unify the SDE framework with causal models. Diff-SCM models the dynamics of causal variables as an Ito process $\mathbf{x}^{(k)}_t, ~ \forall t \in \left[ 0,T \right]$ \citep{ksendal2003StochasticApplications,Sarkka2019AppliedEquations} going from an observed endogenous variable $\mathbf{x}^{(k)}_0 = \mathbf{x}^{(k)}$ to its respective exogenous noise $\mathbf{x}^{(k)}_T = \mathbf{u}^{(k)}$ and back. In other words, we formulate the \textit{forward diffusion as a gradual weakening of the causal relations between variables of a SCM}, as illustrated in Fig.\ \ref{fig:diffusion_causal_relations}. 

The diffusion forces the exogenous noise $\mathbf{u}^{(j)}$ corresponding to a variable $\mathbf{x}^{(j)}$ of interest to be independent of other $\mathbf{u}^{(i)},~ \forall ~ i \neq j$, following the constraints from Eq.\ \ref{eq:u_independence}. The Brownian motion (diffusion) leads to a Gaussian distribution, which can be seen as a prior. Analogously, the original joint distribution entailed by the SCM $p_{\mathfrak{G}}(X)$ diffuses to independent Gaussian distributions equivalent to $p(U)$. As such, the time-dependent joint distribution $p(X_t), ~ \forall t \in \left[ 0,T \right]$ have as bounds $p(X_T) = p(U)$ and $p(\mathbf{x}_0) = p_{\mathfrak{G}}(X)$.  Note that $p(X_t)$ refers to time-dependent distribution over all causal variables $\mathbf{x}^{(k)}$.

We follow \citet{Song2021Score-BasedEquations} in defining the diffusion process from Sec.\ \ref{sec:energy_based_models} in terms of an SDE. Since SDEs are stochastic processes, their solution follows a certain probability distribution instead of a deterministic value. By constraining this distribution to be the same as the distribution $p_{\mathfrak{G}}(X)$ entailed by an SCM $\mathfrak{G}$, we can define a deep structural causal model (DSCM) as a set of SDEs (one for each node $k$):
\begin{equation}
\begin{split}
    & d\mathbf{x}^{(k)} = - \frac{1}{2}\beta_t \mathbf{x}^{(k)} dt +  \sqrt{\beta_t} ~d\mathbf{w}, ~~  \forall k \in [1,K], \\ 
    & \text{where} ~~ p(\mathbf{x}^{(k)}_0) = \prod_{j = k}^K p(\mathbf{x}^{(j)} \mid \mathbf{pa}^{(j)}) ~\text{and}~ p(\mathbf{x}^{(k)}_T) = p(\mathbf{u}^{(k)}).
\end{split}
\end{equation}
Here, $\mathbf{w}$ denotes the Wiener process (or Brownian motion). The first part of the SDE ($- \frac{1}{2}\beta_t \mathbf{x}^{(k)}$) is known as drift function \citep{Sarkka2019AppliedEquations}\footnote{The drift function can potentially be used to define temporal relations between variables as in \citet{Rubenstein2018FromModels} and \citet{Blom2020BeyondModels}.}.

The generative process is the solution of the reverse-time SDE from Eq.\ \ref{eq:reverse_sde} in time. This process is done by iteratively updating the exogenous noise $\mathbf{x}^{(k)}_T = \mathbf{u}^{(k)}$ with the gradient of the data distribution w.r.t. the input variable $\nabla_{\mathbf{x}^{(k)}_t} \log p(\mathbf{x}^{(k)}_t)$, until it becomes $\mathbf{x}^{(k)}_0 = \mathbf{x}^{(k)}$ with:
\begin{equation}
\label{eq:reverse_sde}
d\mathbf{x}^{(k)} = \left[ - \frac{1}{2}\beta_t + \beta_t~ \nabla_{\mathbf{x}^{(k)}_t} \log p(\mathbf{x}^{(k)}_t) \right] dt + \sqrt{\beta_t} \bar{\mathbf{w}}.
\end{equation}

The reverse SDE can, therefore, be considered as the process of strengthening causal relations between variables. More importantly, the iterative fashion of the generative process (reverse SDE) is ideal in a causal framework due to the flexibility of applying interventions. We refer the reader to \citet{Song2021Score-BasedEquations} for a detailed description and proofs of SDE formulation for score-based diffusion models.

\subsection{How to Apply Interventions with Anti-Causal Predictors?}
\label{sec:anti_causal_interventions}

An interesting result of Eq.\ \ref{eq:reverse_sde} is that one only needs the gradients of the distribution entailed by the SCM $p_{\mathfrak{G}}$ for sampling. This allows learning of the anti-causal conditional distributions $p_{\mathfrak{G}^-}$ and applying interventions with the causal mechanism. This can be useful when anti-causal learning is more straightforward \citep{Scholkopf2012OnLearning}. In these cases, one would train classifiers in the anti-causal direction for each edge and diffusion models for each node (over which one wants to measure the effect of interventions) in the graph. Then, one might use the gradients of the classifiers and diffusion models to propagate the intervention in the causal direction over the nodes. Following this idea, proposition \ref{th:interventional_gradients} arises as a result of Eq.\ \ref{eq:reverse_sde}.

%In fact, most machine learning tasks are anti-causal \citep{Kilbertus2018GeneralizationLearning}, \textit{e.g.}\ predicting the class of an image.  The exogenous noise would follow a Gaussian prior in case sampling from the interventional distribution or estimated via forward diffusion as detailed in Sec.\ \ref{sec:counterfactuals}.

\begin{proposition}[Interventions as anti-causal gradient updates]
\label{th:interventional_gradients}
We consider the SCM $\mathfrak{G}$ and a variable $\mathbf{x}^{(j)} \in \mathbf{an}^{(k)}$. The effect observed on $\mathbf{x}^{(k)}$ caused by an intervention on $\mathbf{x}^{(j)}$,
$ p_{\mathfrak{G}}(\mathbf{x}^{(k)} \mid do(\mathbf{x}^{(j)} = x^{(j)}))$, is equivalent to solving a reverse-diffusion process for $\mathbf{x}^{(k)}_t$. Since the sampling process involves taking into account the distribution entailed by $\mathfrak{G}$, it is guided by the gradient of an \textbf{anti-causal} predictor w.r.t. the effect when the cause is assigned a specific value:
\begin{equation}
\label{eq:prop_equation}
    \nabla_{x^{(k)}_t} p_{\mathfrak{G}^-}(\mathbf{x}^{(j)} = x^{(j)} \mid x^{(k)}_t).
\end{equation}

\end{proposition}

Proposition \ref{th:interventional_gradients} respects the principle of independent causal mechanisms (ICM)\footnote{The principle states that ``The causal generative process of a system’s variables is composed of autonomous modules that do not inform or influence each other.''} \citep{Peters2017ElementsInference,Scholkopf2012OnLearning}. It implies independence between the cause distribution and the mechanism producing the effect distribution. As shown in Eq.\ \ref{eq:prop_equation}, sampling with the causal mechanism does not require the distribution of the cause $p(\mathbf{x}^{(j)})$ \citep{Scholkopf2021TowardLearning}.

%The ability to reason under interventions is at the center of causal modeling \citep{Pearl2009causality, Scholkopf2021TowardLearning}. It allows not only sampling from an interventional distribution, but also the estimation of counterfactuals. 

\subsection{Counterfactual Estimation with Diff-SCM}
\label{sec:counterfactuals}
A powerful consequence of building causal models, following \textit{Pearl's Causal Hierarchy}, is the estimation of counterfactuals. Counterfactuals are hypothetical scenarios for a given factual observation under a local intervention. Estimation of counterfactuals differentiates of sampling from an interventional distribution because the changes are applied for a given observation. As detailed in \cite{pearl2016causal}, sec.\ 4.2.4, counterfactual estimation requires three steps:
\begin{choices}
    \item abduction of exogenous noise -- forward diffusion with DDIM algorithm \citep{Song2021DenoisingModels} following Alg.\ \ref{alg:ddim-reverse-sampling} in Appendix \ref{app:ddim};
    \item action -- graph mutilation by erasing the edges between the intervened variable and its parents;
    \item prediction -- reverse diffusion controlled by the gradients of an anti-causal classifier.
\end{choices}

Here, we are interested in estimating $x^{(k)}_{\text{CF}}$ based on the observed (factual) $x_{\text{F}}^{(k)}$ for the random variable $\mathbf{x}^{(k)}$ after assigning a value $x_{\text{CF}}^{(j)}$ to $\mathbf{x}^{(j)} \in \mathbf{an}^{(k)}$, \textit{i.e.}\ applying an intervention $do(\mathbf{x}^{(j)} = x^{(j)}_\text{CF})$. It's equivalent to sample from counterfactual distribution $p_{\mathfrak{G}}(\mathbf{x}^{(k)} \mid do(\mathbf{x}^{(j)} = x^{(j)}_\text{CF}); \mathbf{x}^{(k)} = x^{(k)}_{\text{F}} )$. We will consider a setting where only $\mathbf{x}^{(j)}$ and $\mathbf{x}^{(k)}$ are present in the graph as a simplifying assumption for Alg.\ \ref{alg:counterfactual_estimation}. Considering only two variables removes the need for the graph mutilation explained above. It is also the setting used in our experiments. We will leave an extension to more complex SCMs for future work. We detail in Alg.\ \ref{alg:counterfactual_estimation} how abduction of exogenous noise and prediction is done.

\textbf{Abduction of Exogenous Noise.}~ The first step for estimating a counterfactual is the abduction of exogenous noise. Note from Eq.\ \ref{eq:structural_assignments} that the value of a causal variable depends both on its parents and on its respective exogenous noise. From a deep learning perspective \citep{Pawlowski2020DeepInference}, one might consider the exogenous $\mathbf{u}^{(k)}$ an inferred latent variable. The prior $p(\mathbf{u}^{(k)})$ of $\mathbf{u}^{(k)}$ in Diff-SCM is a Gaussian as detailed in Sec.\ \ref{sec:causal_models_diffusion}.

With diffusion models, abduction can be done with a derivation done by \citet{Song2021DenoisingModels} and \citet{Song2021Score-BasedEquations}. Both works make a connection between diffusion models and neural ODEs \citep{Chen2018NeuralEquations}. They show that one can obtain a deterministic inference system while training with a diffusion process, which is stochastic by nature. This formulation allows the process to be invertible by recovering a latent space $u^{(k)}$ by performing the forward diffusion with the learned model. The algorithm for recovering $u^{(k)}$ is highlighted as the first box in Alg.\ \ref{alg:counterfactual_estimation}.

\textbf{Prediction under Intervention.}~Once the abduction of exogenous noise $u^{(k)}$ is done for a given factual observation $x_{\text{F}}^{(k)}$, counterfactual estimation consists in applying an intervention in the reverse diffusion process with the gradients of an anti-causal predictor. In particular, we use the formulation of guided DDIM from \citet{Dhariwal2021DiffusionSynthesis} which forms the second part of Alg.\ \ref{alg:counterfactual_estimation}.

\textbf{Controlling the Intervention.}~There are three main factors contributing for the counterfactual estimation in Alg.\ \ref{alg:counterfactual_estimation}:
\begin{choices}
    \item The inferred $u^{(k)}$ keeps information about the factual observation;
    \item $\nabla_{x^{(k)}_t} \log p_{\phi}(x_{\text{CF}}^{(j)} \mid x^{(k)}_t)$ guide the intervention towards the desired counterfactual class; and 
    \item $\boldsymbol{\epsilon}_{\theta}(x^{(k)}_t,t) $ forces the estimation to belong to the data distribution.
    
\end{choices}
We follow \citet{Dhariwal2021DiffusionSynthesis} in adding an hyperparameter $s$ which controls the scale of $\nabla_{x^{(k)}_t} \log p_{\phi}(x_{\text{CF}}^{(j)} \mid x^{(k)}_t)$. High values of $s$ might result in counterfactuals that are too different from the factual data. We show this empirically and discuss the effects of this hyperparameter in Sec.\ \ref{sec:hyper_search}.

\begin{algorithm}[ht]
\SetKwInOut{Models}{Models}
\SetKwInOut{Input}{Input}
\SetKwInOut{Output}{Output}
\SetAlgoLined
\Models{trained diffusion model $\boldsymbol{\epsilon}_{\theta}$ and anti-causal predictor $p_{\phi}(x^{(j)} \mid x^{(k)}_t)$}
\Input{factual variable $x_{0,\text{F}}^{(k)}$, target intervention $x_{0,\text{CF}}^{(j)}$, scale $s$}
\Output{counterfactual $x_{0,\text{CF}}^{(k)}$}
\begin{tcolorbox}[colback=blue!5!white,colframe=blue!50!black!50!,left=2pt,right=2pt,top=0pt,bottom=1pt,colbacktitle=blue!25!white,title=\textbf{\color{black} Abduction of Exogenous Noise --  Recovering $u^{(k)}$ from $x_{0,\text{F}}^{(k)}$}]
\For{$t \leftarrow 0$ \KwTo $T$}{
$x^{(k)}_{t+1,\text{F}} \leftarrow \sqrt{\alpha_{t+1}} \left( \frac{x^{(k)}_{t,\text{F}} - \sqrt{1 - \alpha_t} ~ \boldsymbol{\epsilon}_{\theta}(x^{(k)}_{t,\text{F}},t)}{\sqrt{\alpha_t}}  \right) + \sqrt{\alpha_{t+1}} ~ \boldsymbol{\epsilon}_{\theta}(x^{(k)}_{t,\text{F}},t)$
}
$u^{(k)} = x^{(k)}_{T,\text{F}} = x^{(k)}_{T}$
\end{tcolorbox}
\begin{tcolorbox}[colback=teal!5!white,colframe=teal!50!black!50!,left=2pt,right=2pt,top=0pt,bottom=1pt,
  colbacktitle=teal!25!white,title=\textbf{\color{black} Generation under Intervention}]
\For{$t \leftarrow T$ \KwTo $0$}{

$\epsilon \leftarrow \boldsymbol{\epsilon}_{\theta}(x^{(k)}_t,t) - s  \sqrt{1 - \alpha_t} ~ \nabla_{x^{(k)}_t} \log p_{\phi}(x_{0,\text{CF}}^{(j)} \mid x^{(k)}_t)$

$x^{(k)}_{t-1} \leftarrow \sqrt{\alpha_{t-1}} \left( \frac{x^{(k)}_t - \sqrt{1 - \alpha_t} ~ \epsilon}{\sqrt{\alpha_t}}  \right) + \sqrt{\alpha_{t-1}} ~ \epsilon$
}
$x_{0,\text{CF}}^{(k)} = x^{(k)}_0$
\end{tcolorbox}
\caption{Inference of \textbf{counterfactual} for a variable $\mathbf{x}^{(k)}$ from an intervention on $\mathbf{x}^{(j)} \in \mathbf{an}^{(k)}$} \label{alg:counterfactual_estimation}
\end{algorithm}

\section{Related Work}
\textbf{Generative EBMs.}~Our generative framework is inspired on the energy based models literature \citep{Ho2020DenoisingModels,Song2021Score-BasedEquations,Du2019ImplicitModels,Grathwohl2020YourOne}. In particular, we leverage the theory around denoising diffusion models \citep{Sohl-Dickstein2015DeepThermodynamics,Ho2020DenoisingModels,Nichol2021ImprovedModels}. We take advantage of a non-Markovian definition DDIM \citep{Song2021DenoisingModels} which allows faster sampling and recovering latent spaces from observations. Our theory connecting diffusion models and SDEs follows \citet{Song2021Score-BasedEquations}, but from a different perspective. Even though \citet{Du2020CompositionalModelsb} are not constrained to causal modeling, they also use the idea of guiding the generation with gradient of conditional energy models. Recently, \citet{Sinha2021D2C:Generation} proposed a version of diffusion models for manipulable generation based on contrastive learning. Finally, \citet{Dhariwal2021DiffusionSynthesis} derive a conditional sampling process for DDIM that is used in this paper as detailed in Sec.\ \ref{sec:anti_causal_interventions}. Here, we re-interpret their generation algorithm from a causal perspective and add deterministic latent inference for counterfactual estimation. The main, but key difference, is that we add the \textit{abduction of exogenous noise}. Without this abduction, we cannot ensure that the resulting image will match other aspects of the original image whilst altering only the intended aspect (ie. Where we want to intervene). We can sample from a counterfactual distribution instead of the interventional distribution.

\textbf{Counterfactuals.}~Designing causal models with deep learning components has allowed causal inference with high-dimensional variables \citep{Pawlowski2020DeepInference,Shen2020DisentangledLearning, Dash2020EvaluatingCounterfactuals,Xia2021TheInference,Zecevi2021RelatingModels}. Given a factual observation, counterfactuals are obtained by measuring the effect of an intervention in one of the ancestral attributes. They have been used in a range of applications such as
\begin{choices}
     \item explaining predictions \citep{Verma2020CounterfactualReview,Goyal2019CounterfactualExplanations,Looveren2021InterpretablePrototypes,Hvilshj2021ECINN:Networks};
     \item defining fairness \citep{Kusner2017CounterfactualFairness};
     \item mitigating data biases \citep{Denton2019ImageBias};
     \item improving reinforcement learning \citep{Lu2020Sample-EfficientAugmentation};
     \item predicting accuracy \citep{Kaushik2020EXPLAININGDATA};
     \item increasing robustness against spurious correlations \citep{Sauer2021CounterfactualNetworks}.
\end{choices}
Most similar to our work, \citet{Schut2021GeneratingUncertainties} estimate counterfactuals via iterative updates using the gradients of a classifier. However, their method is based on adversarial updates computed via epistemic uncertainty, not diffusion processes. 

\section{Experiments}

Ground truth counterfactuals are, by definition, impossible to acquire. Counterfactuals are hypothetical predictions. In an ideal scenario, the SCM of problem is fully specified. In this case, one would be able to verify if unrelated causal variables kept their values\footnote{Remember that interventions only change descendants in a causal graph.}. However, a complete causal graph is rarely known in practice.
In this section, we 
\begin{choices}
     \item present ideas on how to evaluate counterfactuals without access to the complete causal graph nor semi-synthetic data;
     \item show with quantitative and qualitative experiments that our method is appropriate for counterfactual estimation;
     \item propose CLD, a metric for quantitative evaluation of counterfactuals; and
     \item use CLD for fine tuning an important hyperparameter of our framework.
\end{choices}

\textbf{Causal Setup.}~We consider a causal model $\mathfrak{G}_{image}$ with two variables $\mathbf{x}^{(1)} \leftarrow \mathbf{x}^{(2)}$ following the example in Sec.\ \ref{sec:anti_causal_interventions}. Here, $\mathbf{x}^{(1)}$ represents an image and $\mathbf{x}^{(2)}$ a class. In practice, the gradient of the marginal distribution of $\mathbf{x}^{(1)}$ is learned with a diffusion model, which we refer as $\boldsymbol{\epsilon}_{\theta}$, as in Sec.\ \ref{sec:energy_based_models}. The anti-causal conditional distribution is also learned with a neural network $p_{\phi}(\mathbf{x}^{(2)} \mid \mathbf{x}^{(1)})$. Our experiments aim at sampling from the counterfactual distribution $p_{\mathfrak{G}}(\mathbf{x}^{(1)} \mid do(\mathbf{x}^{(2)} = x^{(2)}_{\text{CF}}); x^{(1)}_{\text{F}})$. Extra experiments on sampling from interventional distribution are in Appendix \ref{app:interventional_experiments}.

\textbf{Implementation.}~ $\boldsymbol{\epsilon}_{\theta}$ is implemented as an encoder-decoder architecture with skip-connections, \textit{i.e.}\ a Unet-like network \citep{Ronneberger2015U-Net:Segmentation}. For anti-causal classification tasks, we use the encoder of $\boldsymbol{\epsilon}_{\theta}$ with a pooling layer followed by a linear classifier. Both $\boldsymbol{\epsilon}_{\theta}$ and $p_{\phi}(\mathbf{x}^{(2)} \mid \mathbf{x}^{(1)})$ dependent on diffusion time. The diffusion model and anti-causal predictor are trained separately. Implementation details are in Appendix \ref{app:implementation_details}. 

\textbf{Baselines.}~We consider \citet{Schut2021GeneratingUncertainties} and \citet{Looveren2021InterpretablePrototypes} because they
\begin{choices}
    \item generate counterfactuals based on classifiers decisions; and
    \item evaluate results with metrics tailored to counterfactual estimation on images.
\end{choices}

\textbf{Datasets.}~Considering the causal model $\mathfrak{G}_{image}$ described above, we compare our method quantitatively and qualitatively with baselines on MNIST data \citep{Lecun1998Gradient-basedRecognition}. Furthermore, we show empirically that our approach works with more complex, higher-resolution images from the ImageNet dataset \citep{Deng2009Imagenet:Database}. We only perform qualitative evaluations on ImageNet since the baseline methods cannot generate counterfactuals for this dataset.

\subsection{Evaluating Counterfactuals: Realism and Closeness to Data Manifold}
\label{sec:evaluation_ims}

Taking into account the causal model $\mathfrak{G}_{image}$, we now employ the strategies for counterfactual estimation in Sec.\ \ref{sec:counterfactuals}. In particular, given an image $x^{(1)}_\text{F} \sim \textbf{x}^{(1)}$ and a target intervention $x_{\text{CF}}^{(2)}$ in the class variable, we wish to estimate the counterfactual $x^{(1)}_\text{CF}$ for the image $x^{(1)}_\text{F}$. We use two metrics proposed by \citet{Looveren2021InterpretablePrototypes}, IM1 and IM2, to measure the realism, interpretability and closeness to the data manifold based on the reconstruction loss of autoencoders trained on specific classes. See details in Appendix \ref{app:ims}.

\textbf{Experimental Setup.}~We run Alg.\ \ref{alg:counterfactual_estimation} over the test set with randomly sampled target counterfactual classes $x^{(2)}_{\text{CF}} \sim \mathbf{x}^{(2)}, ~ \forall x^{(2)} \neq x^{(2)}_{\text{F}}$. For example. we generate counterfactuals of all MNIST classes for a given factual image, as illustrated in Appendix \ref{app:mnist_qualitative}. We evaluate realism of Diff-SCM, \citeauthor{Schut2021GeneratingUncertainties} and \citeauthor{Looveren2021InterpretablePrototypes} using the IM1 and IM2 metrics. Diff-SCM achieves better results (lower is better) in both metrics\footnote{We highlight that our setting is slightly different from baseline works where the target counterfactual classes were similar to the factual classes. \textit{e.g.}\ Transforming MNIST digits from $2 \rightarrow [3,7]$ or $4 \rightarrow [1, 9]$. Since we are sampling target classes randomly, their metric values will look lower than in their respective papers.}, as shown in Tab.\ \ref{tab:quantitative_comparison}. We show qualitative results on ImageNet in Fig.\ \ref{fig:qualitative_ImageNet_counterfactuals_samples} and on MNIST in Appendix \ref{app:mnist_qualitative}. A qualitative comparison between methods is depicted in Fig.\ \ref{fig:qualitative_comparison}.

\begin{table}[hb]
    \caption{Quantitative comparison between Diff-SCM and baselines. Lower is better for all metrics. Results are presented with mean ($\mu$) and standard deviation $\sigma$ over the test set in the format $\mu_{\sigma}$.}
    \label{tab:quantitative_comparison}
    \centering
    \begin{tabular}{c||c||c||c}
Method             & IM1 $\downarrow$ & IM2 $\downarrow$ & CLD $\downarrow$ \\ \hline
Diff-SCM (ours)    &  $\boldsymbol{0.94_{0.02}}$   &   $\boldsymbol{0.04_{0.00}}$  &  $\boldsymbol{1.08_{0.03}}$
\\
\citeauthor{Looveren2021InterpretablePrototypes} &  $1.10_{0.03}$   &   $0.05_{0.00}$   &  $1.25_{0.03}$
\\
\citeauthor{Schut2021GeneratingUncertainties}       &  $1.05_{0.01}$   &   $0.10_{0.00}$   &  $ 1.19_{0.01}$ 

    \end{tabular}
\end{table}

\subsection{Counterfactual Latent Divergence (CLD)}
\label{sec:cld}

\begin{figure}[ht]
\floatconts
{fig:example2}% label for whole figure
{\caption{(\emph{a}) A t-SNE visualization of the 20-dimensional latent vector of a variational autoencoder VAE over all MNIST samples. Each point represents an MNIST image and colors represent the ground-truth label of each sample. CLD's goal is to estimate a relative similarity between the factual data and the counterfactual. The distance between the generated counterfactual $do(0)$ and factual observation is compared to the distances between the factual observation and all other data points from factual and counterfactual classes. (\emph{b}) Qualitative comparison with baselines approaches for counterfactual estimation. Each column represents one method and each row a different intervention on digit class. The \emph{train.}\ column shows training samples belonging to the target intervention class.}}% caption for whole figure
{%
\subfigure[CLD Intuition]{%
\label{fig:cld_intuition}% label for this sub-figure
\includegraphics{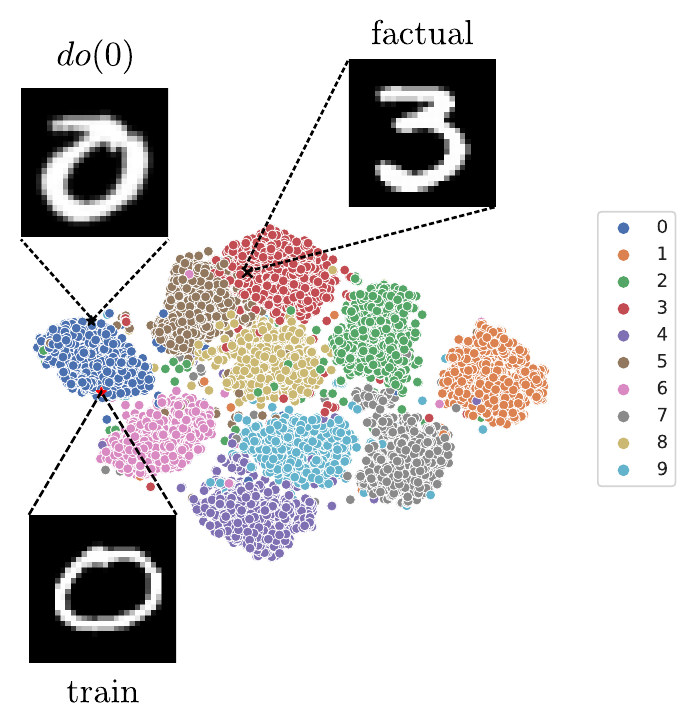}%
}\qquad % space out the images a bit
\subfigure[Qualitative comparison]{%
\label{fig:qualitative_comparison}% label for this sub-figure
\includegraphics{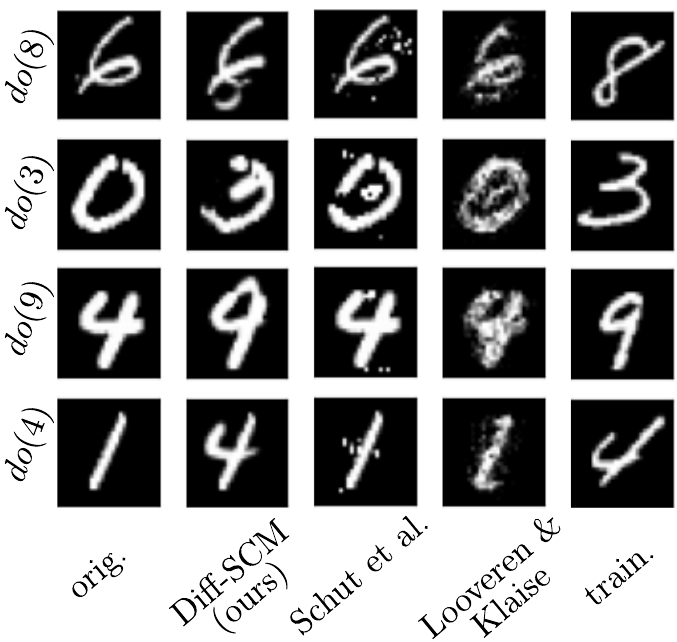}
}
}
\end{figure}

Since one cannot measure changes in all variables of a real SCM, we leverage the sparse mechanism shift (SMS) hypothesis\footnote{SMS states that a ``small distribution changes tend to manifest themselves in a sparse or local way in the causal factorization, that is, they should usually not affect all factors simultaneously."} \citep{Scholkopf2021TowardLearning} for justifying a \textit{minimality} property of counterfactuals. SMS translates, in our setting, to \textit{an intervention will not change many elements of the observed data}. Therefore, an important property of counterfactuals is minimality or proximity to the factual observation. We suggest here a new metric entitled counterfactual latent divergence (CLD), illustrated in Fig.\ \ref{fig:cld_intuition}, that estimates minimality.

Note that the metrics IM1 and IM2 from Sec.\ \ref{sec:evaluation_ims} do not take minimality into account. In addition, previous work \citep{Wachter2018CounterfactualGDPR,Schut2021GeneratingUncertainties} only used the mean absolute error or $\ell_1$ distance in the data space for measuring minimality. However, measuring similarity at pixel-level can be challenging as an intervention might change the structure of the image whilst keeping other factors unchanged. In this case, a pixel-level comparison might not be informative about the other factors of variation. %

\textbf{Latent Similarity.}~Therefore, we choose to measure similarity between latent representation. In addition, we want a representation that captures all factors of variation on the input data.
In particular, we train a variational autoencoder (VAE) \citep{Kingma2014Auto-EncodingBayes} for recovering probabilistic latent representations that capture all factors of variation in the data. The latent spaces computed with the VAE's encoder $E_{\phi}$ are denoted as $\mu_i , \sigma_i = E_{\phi}(x^{(1)}_i)$, where subscript $i$ means different samples from $\mathbf{x}^{(1)}$ ($t = 0$). We use the Kullback–Leibler divergence (KL) divergence for measuring the distances between latents. %Since the representations are parametrized as a multivariate Gaussians, the KL divergence (Eq.\ \ref{eq:kl_div}) has as closed form. 
The divergence for a given counterfactual estimation and factual observation pair $(x^{(1)}_{\text{CF}},x^{(1)}_{\text{F}})$ can, therefore, be denoted as
\begin{equation}
    \label{eq:kl_div}
    div = D(x^{(1)}_{\text{CF}},x^{(1)}_{\text{F}}), \qquad  \text{with} ~~  D(x^{(1)}_i,x^{(1)}_j) = \displaystyle D_{\text{KL}}(\mathcal{N}(\mu_i,\sigma_i),\mathcal{N}(\mu_j,\sigma_j)).
\end{equation}

\textbf{Relative Measure.}~However, absolute similarity measures give limited information. Therefore, we leverage class information for measuring minimality whilst making sure that the counterfactual is far enough from the factual class. A relative measure is obtained by estimating the probability of sets of divergence measures between the factual observation and other data points in the dataset (formalized in the Eq.\ \ref{eq:divergence_sets}) to less or greater than $div$. In particular, we compare $div$ with the set $\set{S}_\text{class}$ of divergence measures between the factual observation $x^{(1)}_{\text{F}}$ and all data points $x^{(1)}$ in a dataset $\mathcal{D} = \{(x^{(1)},x^{(2)}) \mid x^{(1)} \in \R^2 ,  x^{(2)} \in \mathbb{N} \}$ for which the class $x^{(2)}$ is $x^{(2)}_\text{class}$ is denoted in set-builder notation\footnote{ We use the following set-builder notation: $\text{MY\_SET} = \{ \text{function}(\text{input}) \mid  \text{input\_domain}\}$.} with:
\begin{equation}
\label{eq:divergence_sets}
    \set{S}_\text{class} = \{ D(x^{(1)},x^{(1)}_{\text{F}})  \mid  (x^{(1)},x^{(2)}) \in \mathcal{D} ~ \land ~ x^{(2)} =  x^{(2)}_\text{class} \}.
\end{equation}
The sets $\set{S}_\text{CF}$ and $\set{S}_\text{F}$ are obtained by replacing ``class" in $\set{S}_\text{class}$ with the appropriate target class of the counterfactual and factual observation class respectively. 

The relative measures are: 
\begin{choices}
    \item $P(\set{S}_\text{CF} \leq div)$ for comparing $div$ with the distance between all data points of the counterfactual class and the factual image; and
    \item $P(\set{S}_\text{F} \geq div)$ for comparing $div$ with the distance between all other data points of the factual class and the factual image. 
\end{choices}
We aim for counterfactuals with low $P(\set{S}_\text{CF} \leq div)$, enforcing minimality, and low $P(\set{S}_\text{F} \geq div)$, enforcing bigger distances from the factual class.

\textbf{CLD.}~We highlight the competing nature of the two measures $P(\set{S}_\text{CF} \leq div)$ and $P(\set{S}_\text{F} \geq div)$ in the counterfactual setting. For example, if the intervention is too minimal -- \textit{i.e.}\ low $P(\set{S}_\text{CF} \leq div)$ -- the counterfactual will still resemble observations from the factual class -- \textit{i.e.}\ high $P(\set{S}_\text{F} \geq div)$. Therefore, the goal is to find the best balance between the two measures. Finally, we define the counterfactual latent divergence (CLD) metric as the LogSumExp of the two probability measures. The LogSumExp operation acts as a smooth approximation of the maximum function. It also penalizes relative peak values for any of the measures when compared to a simple summation. We denote CLD as:
\begin{equation}
    \text{CLD} = \log \left( \exp \left( P\left( \set{S}_\text{CF} \leq div \right) \right) + \exp \left( P\left( \set{S}_\text{F} \geq div \right) \right) \right).
\end{equation}
We show, using the same experimental setup as in Sec.\ \ref{sec:evaluation_ims}, that CLD improves counterfactual estimation when quantitatively compared with the baseline methods, as illustrated in Tab.\ \ref{tab:quantitative_comparison}.

\subsection{Tuning the Hyperparameter $s$ with CLD}
\label{sec:hyper_search}

We now utilize CLD, the proposed metric, for fine-tuning $s$, the scale hyperparameter of our framework detailed in Sec.\ \ref{sec:counterfactuals}. Incidentally, the model with hyperparameters achieving best CLD outperforms previous methods in other metrics (see Tab.\ \ref{tab:quantitative_comparison}) and output the best qualitative results (see Fig.\ \ref{fig:qualitative_comparison}). This result further validate that our metric is suited for counterfactual evaluation.

\begin{wrapfigure}{lt}{0.5\textwidth}
  \centering
    \includegraphics[width=0.5\textwidth]{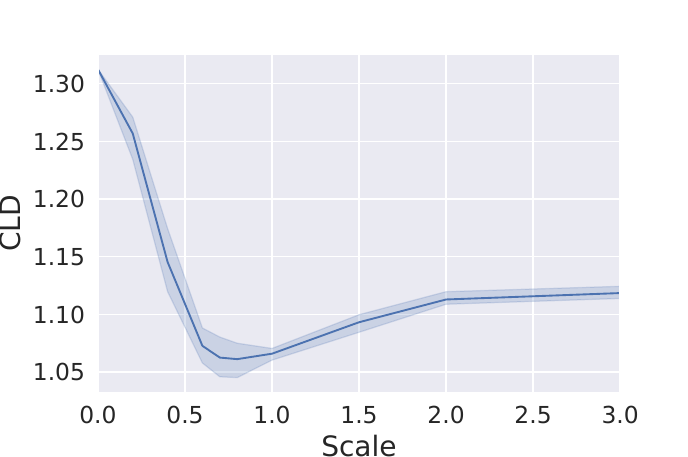}
  \caption{Scale hyperparameter search using CLD (lower is better). The line plot shows the mean and 95\% confidence interval. We found that $s = 0.7$ is the best value.}
  \label{fig:cld_vs_scale}
  %\vspace{-0.5cm}
\end{wrapfigure}

\textbf{Experimental Setup.}~We run Alg.\ \ref{alg:counterfactual_estimation} while varying the scale hyperparameter $s$ in the $\left[0.0,3.0\right]$ interval for MNIST data, as depicted in Fig.\ \ref{fig:cld_vs_scale}. When $s=0$, the classifier does not influence the generation, therefore, the counterfactuals are reconstructions of the factual data; resulting in a high CLD.

When $s = 3$ (too high), the diffusion model contributes much less than the classifier, therefore, the counterfactuals are driven towards the desired class while ignoring the exogenous noise of a given observation. High values of $s$ correspond to strong interventions which do not hold the minimality property, also resulting in a high CLD. Therefore, the optimum point for $s$ is an intermediate value where CLD is minimum. All MNIST experiments were performed using $s = 0.7$, following this hyperparameter search. See Appendix \ref{app:classifier_scale} for qualitative results.

\section{Conclusions}

We propose a theoretical framework for causal estimation using generative diffusion models entitled Diff-SCM. Diff-SCM unifies recent advances in generative energy-based models and structural causal models. Our key idea is to use gradients of the marginal and conditional distributions entailed by an SCM for causal estimation. The main benefit of only using the distribution's gradients is that one can learn an anti-causal mechanism and use its gradients as a causal mechanism for generation. We show empirically how it can be applied to a two variable causal model. We leave the extension to more complex causal models to future work.

Furthermore, we present an algorithm for performing interventions and estimating counterfactuals with Diff-SCM. We acknowledge the difficulty of evaluating counterfactuals and propose a metric entitled counterfactual latent divergence (CLD). CLD measures the distance, in a latent space, between the observation and the generated counterfactual by comparison with other distances between samples in the dataset. We use CLD for comparison with baseline methods and for hyperparameter search. Finally, we show that the proposed Diff-SCM achieves better quantitative and qualitative results compared to state-of-the-art methods for counterfactual generation on MNIST.

\textbf{Limitations and future work.}~  We only have specifications for two variables in our empirical setting, therefore, applying an intervention on $\mathbf{x}^{(2)}$ means changing all the correlated variables within this dataset. Applying Diff-SCM to more complex causal models would require the use of additional techniques. For instance, consider the SCM depicted in Fig. \ref{fig:diffusion_causal_relations}, a classifier naively trained to predict $x^{(2)}$ (class) from $x^{(1)}$ (image) would be biased towards the confounder $x^{(3)}$. Therefore, the gradient of the classifier w.r.t the image would also be biased. This would make the intervention do($x^{(2)}$) not correct. In this case, the graph mutilation (removing edges from parents of node intervened on) would not happen because the gradients from the classifier would pass information about $x^{(3)}$. We leave this extension for future work.

\section{Acknowledgement}
We thank Spyridon Thermos, Xiao Liu, Jeremy Voisey,  Grzegorz Jacenkow and Alison O'Neil for their input on the manuscript and research support. This work was supported by the University of Edinburgh, the Royal Academy of Engineering and Canon Medical Research Europe via Pedro Sanchez's PhD studentship. This work was partially supported by the Alan Turing Institute under the EPSRC grant EP N510129\textbackslash 1. We thank Nvidia for donating a Titan\-X GPU. S.A. Tsaftaris acknowledges the support of Canon Medical and the Royal Academy of Engineering and the Research Chairs and Senior Research Fellowships scheme (grant RCSRF1819\textbackslash825).

\bibliography{references}
\newpage
\appendix

\section{Theory for Training Diffusion Models}
\label{app:ddpm_definition}

We now review with more detailed the formulation of Denoising Diffusion Probabilistic Models (DDPMs) \citep{Ho2020DenoisingModels}. In DDPM, samples are generated by reversing a diffusion process with a neural network from a Gaussian prior distribution. We begin by defining our data distribution $x_0 \sim p(\mathbf{x}_0)$ and a Markovian noising process which gradually adds noise to the data to produce noised samples $\mathbf{x}_t$ up to $\mathbf{x}_T$. In particular, each step of the noising process adds Gaussian noise according to some variance schedule given by $\beta_t$:
\begin{equation}
    p\left(\mathbf{x}_t \mid \mathbf{x}_{t-1} \right)=\mathcal{N}\left(\mathbf{x}_t ; \sqrt{1-\beta_{t}} ~ \mathbf{x}_{t-1}, \beta_{t} \mathrm{I}\right)
\end{equation}

In addition, it's possible to sample $\mathbf{x}_t$ directly from $\mathbf{x}_0$ without repeatedly sample from $\mathbf{x}_t \sim p\left(\mathbf{x}_t \mid \mathbf{x}_{t-1} \right)$. Instead, $p\left(\mathbf{x}_t \mid \mathbf{x}_{0} \right)$ can be expressed as a Gaussian distribution by defining a variance of the noise for an arbitrary timestep $\alpha_{t}:=\prod_{j=0}^{t}\left(1-\beta_{j}\right)$. We, therefore, proceed to define 
\begin{alignat}{2}
    p\left(\mathbf{x}_t \mid \mathbf{x}_{0} \right) &= \mathcal{N}(\mathbf{x}_t; \sqrt{\alpha}_t \mathbf{x}_0, (1-\alpha_t) \mathrm{I}) \\
    &= \sqrt{\alpha_t} \mathbf{x}_0 + \epsilon \sqrt{1-\alpha_t},\text{  } \epsilon \sim \mathcal{N}(0, \mathrm{I}) \label{eq:jumpnoise}
\end{alignat}

However, we are interested in a generative process which consists in performing a reverse diffusion, going from noise $\mathbf{x}_T$ to data $\mathbf{x}_0$. As such, the model trained with parameters $\theta$ should correspond to conditional distribution $p_{\theta}\left(\mathbf{x}_{t-1} \mid \mathbf{x}_{t} \right)$.

Using Bayes theorem, one finds that the posterior $p(\mathbf{x}_{t-1}|\mathbf{x}_t,\mathbf{x}_0)$ is also a Gaussian with mean $\tilde{\mu}_t(\mathbf{x}_t,\mathbf{x}_0)$ and variance $\tilde{\beta}_t$ defined as follows:

\begin{alignat}{2}
    \tilde{\mu}_t(\mathbf{x}_t,\mathbf{x}_0) \coloneqq
    \frac{\sqrt{\alpha_{t-1} - \alpha_{t}}}{1-\alpha_t}\mathbf{x}_0 + \frac{\alpha_t(1-\alpha_{t-1})}{\alpha_{t-1} (1-\alpha_t)} \mathbf{x}_t \qquad && \qquad
    \tilde{\beta}_t \coloneqq \frac{1-\alpha_{t-1}}{1-\alpha_t} \beta_t 
\end{alignat}

\begin{equation}
    p(\mathbf{x}_{t-1}|\mathbf{x}_t,\mathbf{x}_0) = \mathcal{N}(\mathbf{x}_{t-1}; \tilde{\mu}(\mathbf{x}_t, \mathbf{x}_0), \tilde{\beta}_t \mathrm{I}) \label{eq:posterior}
\end{equation}

Training $p_{\theta}\left(\mathbf{x}_{t-1} \mid \mathbf{x}_{t} \right)$ such that $p(\mathbf{x}_0)$ learns the true data distribution, the following variational lower-bound $L_{\text{vlb}}$ for $p_{\theta}(\mathbf{x}_0)$ can be optimized:
\begin{equation}
    L_{\text{vlb}} \coloneqq -\log p_{\theta}(\mathbf{x}_0 | x_1) + \sum_{t = 2}^{T} \kld{p(\mathbf{x}_{t-1}|\mathbf{x}_t,\mathbf{x}_0)}{p_{\theta}(\mathbf{x}_{t-1}|\mathbf{x}_t)}
\end{equation}

\citet{Ho2020DenoisingModels} considered a variational approximation of the Eq.\ \ref{eq:posterior} for training $p_{\theta}\left(\mathbf{x}_{t-1} \mid \mathbf{x}_{t} \right)$ efficiently. Instead of directly parameterize $\mu_{\theta}(\mathbf{x}_t,t)$ as a neural network, a model $\epsilon_{\theta}(\mathbf{x}_t,t)$ is trained to predict $\epsilon$ from Equation \ref{eq:jumpnoise}. This simplified objective is defined as follows:

\begin{alignat}{2}
    L_{\text{simple}} &\coloneqq \mathbb{E}_{t,\mathbf{x}_0 \sim p_{\text{data}},\epsilon \sim \mathcal{N}\left(0, \mathrm{I}\right)} \left[ (1 - \alpha_t)
\left\| \boldsymbol{\epsilon}_{\theta}(\sqrt{\alpha_t}\mathbf{x}_0 + \sqrt{1 - \alpha_t} \epsilon, t)- \epsilon \right\|_{2}^{2}\right]
\end{alignat}

\section{Pearl's Causal Hierarchy}
\label{app:pch}
\citet{Bareinboim2020OnInference} use \textit{Pearl's Causal Hierarchy} (PCH) nonmenclature after Pearl's seminal work on causality which is well illustrated in \citet{Pearl2018TheEffect} as the \textit{Ladder of Causation}. PCH states that structural causal models should be able to sample from a collection of three distributions (\cite{Peters2017ElementsInference}, Ch. 6) which are related to cognitive capabilities:
\begin{enumerate}
    \item The \textit{observational} (``seeing'') distribution $p_{\mathfrak{G}}(\mathbf{x}^{(k)})$.
    \item The do-calculus \citep{Pearl2009causality} formalizes sampling from the \textit{interventional} (``doing'') distribution $p_{\mathfrak{G}}(\mathbf{x}^{(k)} \mid do(\mathbf{x}^{(j)} = x^{(j)} ))$. The $do()$ operator means an intervention on a specific variable is propagated only through it's descendants in the SCM $\mathfrak{G}$. The causal structure forces that only the descendants of the variable intervened upon will be modified by a given action.
    \item Sampling from a \textit{counterfactual} (``imagining'') distribution $p_{\mathfrak{G}}(\mathbf{x}^{(k)} \mid do(\mathbf{x}^{(j)} = x^{(j)}); x^{(k)} )$ involves applying an intervention $do(\mathbf{x}^{(j)} = x^{(j)})$ on an given instance $\mathbf{x}^{(k)}$ . Contrary to the factual observation, a counterfactual corresponds to a hypothetical scenario.
\end{enumerate}

\section{Example of Anti-causal Intervention}
\label{app:example}
We illustrate Prop. \ref{th:interventional_gradients} in a case with two variables, which is also used in the experiments. Consider a variable $\mathbf{x}^{(1)}$ caused by $\mathbf{x}^{(2)}$, \textit{i.e.}\ $\mathbf{x}^{(1)} \leftarrow \mathbf{x}^{(2)}$. Following the causal direction, the joint distribution can be factorised as $p(\mathbf{x}^{(1)},\mathbf{x}^{(2)}) =p(\mathbf{x}^{(1)} \mid \mathbf{x}^{(2)})p(\mathbf{x}^{(2)})$. Applying an intervention with the SDE framework, however, one would only need $\nabla_{x^{(1)}} \log p_t(\mathbf{x}^{(1)} \mid \mathbf{x}^{(2)} = x^{(2)})$, as in Eq.\ \ref{eq:reverse_sde}. By applying Bayes' rule, one can derive $p(\mathbf{x}^{(1)} \mid \mathbf{x}^{(2)}) =p(\mathbf{x}^{(2)} \mid \mathbf{x}^{(1)})p(\mathbf{x}^{(1)}) /p(\mathbf{x}^{(2)})$.  Therefore, the sampling process would be done with

\begin{equation}
\label{eq:bayes_rule_grad}
    \nabla_{x^{(1)}} \log p(\mathbf{x}^{(1)} \mid \mathbf{x}^{(2)}) \propto \nabla_{x^{(1)}} \log p(\mathbf{x}^{(2)} \mid \mathbf{x}^{(1)}) + \nabla_{x^{(1)}} \log p(\mathbf{x}^{(1)}).
\end{equation}

\section{DDIM sampling procedure}
\label{app:ddim}

A variation of the DDPM~\citep{Ho2020DenoisingModels} sampling procedure is done with Denoising Diffusion Implicit Models (DDIM, \cite{Song2021DenoisingModels}). DDIM formulates an alternative non-Markovian noising process
that allows a deterministic mapping between latents to images. The deterministic mapping means that the noisy term in Eq.\ \ref{eq:ddpm_sampling} is no longer necessary for sampling. This sampling approach has the same forward marginals as DDPM, therefore, it can be trained in the same manner. This approach was used for sampling throughout the paper as explained in Sec.\ \ref{sec:counterfactuals}. 

Alg.\ \ref{alg:ddim-sampling} describes DDIM's sampling procedure from $\mathbf{x}_T \sim \gN(0, \mI)$ (exogenous noise distribution) to $\mathbf{x}_0$ (data distribution) deterministic procedure.
This formulation has two main advantages:
\begin{choices}
    \item it allows a near-invertible mapping between $\mathbf{x}_T$ and $\mathbf{x}_0$ as shown in Alg.\ \ref{alg:ddim-reverse-sampling}; and
    \item it allows efficient sampling with fewer iterations even when trained with the same diffusion discretization. This is done by choosing different undersampling $t$ in the $[0,T]$ interval.
\end{choices}

\begin{algorithm}[ht]
\SetKwInOut{Models}{Models}
\SetKwInOut{Input}{Input}
\SetKwInOut{Output}{Output}
\SetAlgoLined
\Models{trained diffusion model $\boldsymbol{\epsilon}_{\theta}$.}
\Input{$x_{T} \sim \mathcal{N}(0,\mathrm{I})$}
\Output{$x_{0}$ - Image}

\For{$t \leftarrow T$ \KwTo $0$}{

$x_{t-1} \leftarrow \sqrt{\alpha_{t-1}} \left( \frac{x_t - \sqrt{1 - \alpha_t} ~ \boldsymbol{\epsilon}_{\theta}(x_t,t)}{\sqrt{\alpha_t}}  \right) + \sqrt{\alpha_{t-1}} ~ \boldsymbol{\epsilon}_{\theta}(x_t,t)$
}
\caption{Sampling with DDIM - Image Generation} \label{alg:ddim-sampling}
\end{algorithm}

\begin{algorithm}[ht]
\SetKwInOut{Models}{Models}
\SetKwInOut{Input}{Input}
\SetKwInOut{Output}{Output}
\SetAlgoLined
\Models{trained diffusion model $\boldsymbol{\epsilon}_{\theta}$.}
\Input{$x_{0}$ - Image  } 
\Output{$x_{T}$ - Latent Space}

\For{$t \leftarrow T$ \KwTo $0$}{

$x_{t+1} \leftarrow \sqrt{\alpha_{t+1}} \left( \frac{x_t - \sqrt{1 - \alpha_t} ~ \boldsymbol{\epsilon}_{\theta}(x_t,t)}{\sqrt{\alpha_t}}  \right) + \sqrt{\alpha_{t+1}} ~ \boldsymbol{\epsilon}_{\theta}(x_t,t)$
}
\caption{Reverse-Sampling with DDIM - Inferring the Noisy Latent} \label{alg:ddim-reverse-sampling}
\end{algorithm}

\section{Implementation Details}
\label{app:implementation_details}
For each dataset, we train two models that are trained separately:
\begin{choices}
\item $\boldsymbol{\epsilon}_{\theta}$ is implemented as an encoder-decoder architecture with skip-connections, \textit{i.e.}\ a Unet-like network
\citep{Ronneberger2015U-Net:Segmentation}.
\item A (Anti-causal) classifier that uses the encoder of $\boldsymbol{\epsilon}_{\theta}$ with a pooling layer followed by a linear classifier.
\end{choices}
All models are time conditioned. Time, which is a scalar, is embedded using the transformer's sinusoidal position embedding \citep{Vaswani2017AttentionNeed}. The embedding is incorporated into the convolutional models with an Adaptive Group Normalization layer into each residual block \citep{Nichol2021ImprovedModels}. Our architectures and training procedure follow \citet{Dhariwal2021DiffusionSynthesis}. They performed an extensive ablation study of important components from DDPM \citep{Ho2020DenoisingModels} and improved overall image quality and log-likelihoods on many image benchmarks. We use the same hyperparameters as \citet{Dhariwal2021DiffusionSynthesis} for the ImageNet and define ours for MNIST. The specific hyperparameters for diffusion and classification models follow Tab.\ \ref{tab:hpsdiff}. We train all of our models using Adam with $\beta_1=0.9$ and $\beta_2=0.999$. We train in 16-bit precision using loss-scaling, but maintain 32-bit weights, EMA, and optimizer state. We use an EMA rate of 0.9999 for all experiments.

We use DDIM sampling for all experiments with 1000 timesteps. The same noise schedule is used for training. Even though DDIM allows faster sampling, we found that it does not work well for counterfactuals.

\begin{table}[ht]
    \setlength\tabcolsep{4pt}
    \begin{center}
    \begin{small}
    \begin{tabular}{l||cc|cc}
    %\toprule
     dataset & ImageNet 256 & ImageNet 256 & MNIST & MNIST \\
    \midrule
    model & diffusion & classifier & diffusion & classifier \\
    Diffusion steps & 1000 & 1000 & 1000 & 1000 \\
    Model size & 554M  & 54M  & 2M  & 500K\\
    Channels & 256 & 128 & 64 & 32 \\
    Depth &  2 &  2  &  1 &  1\\
    Channels multiple & 1,1,2,2,4,4 & 1,1,2,2,4,4 & 1,2,4 & 1,2,4,4 \\
    Attention resolution &  32,16,8 &  32,16,8 &  -  &  -  \\
    Batch size & 256 & 256 & 256 & 256 \\
    Iterations & $\approx 2$M & $\approx 500$K   & 30K & 3K  \\ 
    Learning Rate & 1e-4  & 3e-4 & 1e-4  & 1e-4\\
    %\bottomrule
    \end{tabular}
    \end{small}
    \end{center}
    \caption{Hyperparameters for models.}
    \label{tab:hpsdiff}
     \vskip -0.2in
\end{table}

\section{Sampling from The Interventional Distribution}
\label{app:interventional_experiments}

In this section, we make sure that our method complies with the second level of \textit{Pearl's Causal Hierarchy} (details in Appendix \ref{app:pch}). Diff-SCM can be used for efficiently sampling from the interventional distributions $p_{\mathfrak{G}_{\text{image}}}(\mathbf{x}^{(1)} \mid do(\mathbf{x}^{(2)} = x^{(2)}))$. Sampling from the interventional distribution can be done by using the second part (``Generation with Intervention'') of Alg.\ \ref{alg:counterfactual_estimation} but sampling $u^{(k)}$ from a Gaussian prior, instead of inferring the latent space (using ``Abduction of Exogenous Noise"). This formulation is identical to \citet{Dhariwal2021DiffusionSynthesis} with guided DDIM \citep{Song2021DenoisingModels} (details in appendix \ref{app:ddim}). \citet{Dhariwal2021DiffusionSynthesis} achieves state-of-the-art image quality results in generation while providing faster sampling than DDPM. Since its capabilities in image synthesis compared to other generative models are shown in \citet{Dhariwal2021DiffusionSynthesis}, we restrict ourselves to present qualitative results on ImageNet 256x256. 

\begin{figure}[h]
\centering
\includegraphics[width=\linewidth]{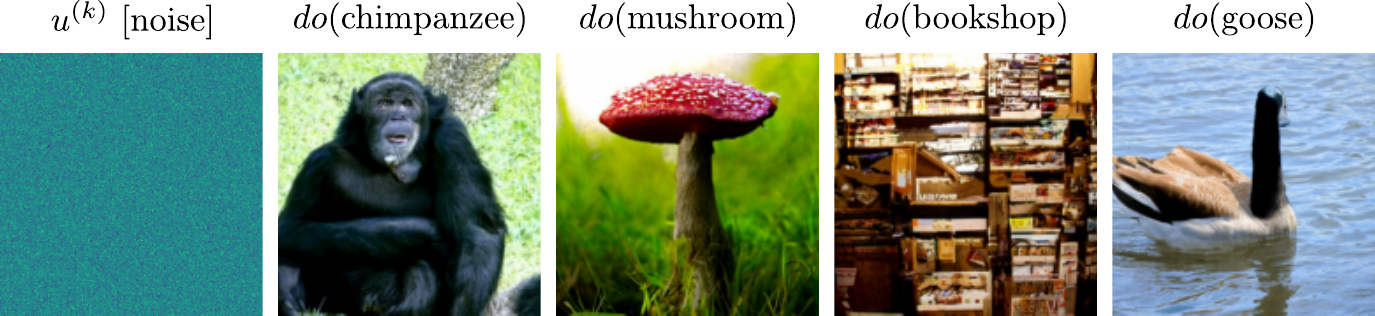}
\caption{Sampling ImageNet images from the interventional distribution. All images originate from the same initial noise $u^{(k)}$ but different interventions are applied at inference time.}
\label{fig:interventional_fig}
\end{figure}

\textbf{Experimental Setup.}~Our experiment, depicted in Fig.\ \ref{fig:interventional_fig}, consists in sampling a single latent space $u^{(1)}$ from a Gaussian distribution and generating samples for different classes. Since all images are generated from the same latent, this allows visualization of the effect of the classifier guidance for different classes. This setup differs from experiments in \citet{Dhariwal2021DiffusionSynthesis}, where each image presented was a different sample $u^{(1)} \sim \textbf{u}^{(1)}$. Here, by sampling $\textbf{u}^{(1)}$ only once, we isolate the contribution of the causal mechanism from the sampling of the exogenous noise $\textbf{u}^{(1)}$. We use the scale hyperparameter $s = 5$ for these experiments.

\section{IM1 and IM2}
\label{app:ims}
\citet{Looveren2021InterpretablePrototypes} propose IM1 and IM2 for measuring the realism and closeness to the data manifold. These metrics are based on the reconstruction losses of auto-encoders trained on specific classes:
\begin{equation}
    \text{IM1}(x^{(1)}_{\text{CF}}, x^{(2)}_{\text{F}}, x_{\text{CF}}^{(2)}) = \frac{\norm{x^{(1)}_{\text{CF}} - \mathrm{AE}_{x_{\text{CF}}^{(2)}}(x^{(1)}_{\text{CF}})}^2_2}{\norm{x^{(1)}_{\text{CF}} - \mathrm{AE}_{x^{(2)}_{\text{F}}}(x^{(1)}_{\text{CF}})}^2_2 + \epsilon}
\end{equation}

\begin{equation}
\label{eq:im1}
    \text{IM2}(x^{(1)}_{\text{CF}}, x_{\text{CF}}^{(2)}) =   \frac{\norm{\mathrm{AE}_{x^{(2)}_{\text{F}}}(x^{(1)}_{\text{CF}}) - \mathrm{AE}(x^{(1)}_{\text{CF}})}_2^2}{\norm{x^{(1)}_{\text{CF}}}_1 + \epsilon}        
\end{equation}

where $\mathrm{AE}_{x^{(2)}}$ denotes an autoencoder trained only on instances from class $x^{(2)}$, and $\mathrm{AE}$ is an autoencoder trained on data from all classes. IM1 is the ratio of the reconstruction loss of an autoencoder trained on the counterfactual class divided by the loss of an autoencoder trained on all classes. IM2 is the normalized difference between the reconstruction of the CF under an autoencoder trained on the counterfactual class, and one trained on all classes.

\section{More MNIST Counterfactuals}
\label{app:mnist_qualitative}

\begin{figure}[h]
\centering
\includegraphics[width=\linewidth]{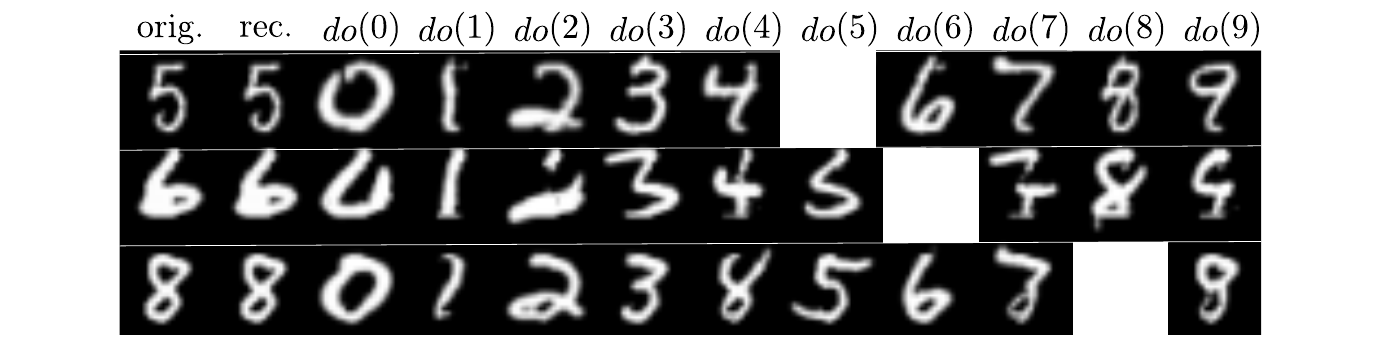}
\caption{MNIST counterfactuals. From the left to right, one can observe the original image (\textit{orig.}), the reconstruction (\textit{rec.}, which entails in running the algorithm \ref{alg:counterfactual_estimation} without the anti-causal predictor) and the resulting counterfactuals for each of the digit classes in the dataset.}
\label{fig:qualitative_MNIST_counterfactuals_samples}
\end{figure}

Here, we show in Fig.\ \ref{fig:qualitative_MNIST_counterfactuals_samples} that we can generate counterfactuals of all MNIST classes, given factual image. We use the scale hyperparameter $s = 0.7$ for these experiments.

\section{Qualitative influence of classifier scale}
\label{app:classifier_scale}

\begin{figure}[h]
\centering
\includegraphics[width=.8\linewidth]{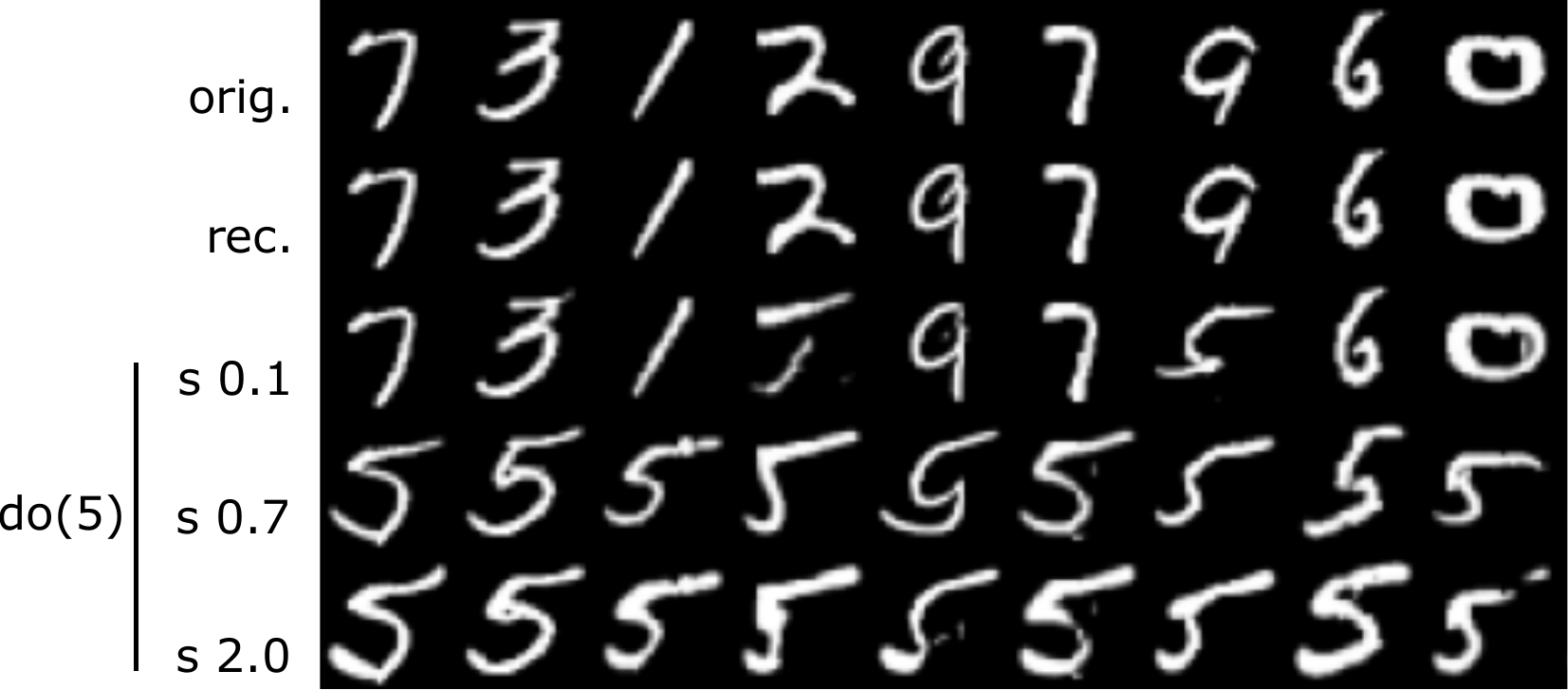}
\caption{MNIST counterfactuals. From top to bottom, one can observe the original image (\textit{orig.}), the reconstruction (\textit{rec.}, and the resulting counterfactuals for the intervention $do(5)$ over three scales. As shown in Fig.\ \ref{fig:cld_vs_scale}, $s = 0.7$ is the optimal scale for MNIST data.}
\label{fig:classifier_scale}
\end{figure}

Here, we show in Fig.\ \ref{fig:classifier_scale} the influence of changing the classifier's scale $s$ quantitatively. If $s$ is too low, the intervention will have a mild effect. On the other had, if $s$ is too high, the intervention will neglect the information present in the exogenous noise, therefore, the counterfactual is maintain less factors from the original image.

\end{document}